\documentclass[manuscript,nonacm]{acmart}
\AtBeginDocument{%
  }

\usepackage[table]{xcolor}
\usepackage{graphicx}
\usepackage{subcaption}
\usepackage{float}
\usepackage{adjustbox}

\begin{document}

\title{CFL: On the Use of Characteristic Function Loss for Domain Alignment in Machine Learning}

\author{Abdullah Almansour}
\authornote{Both authors contributed equally to this research.}
\email{aalmanso@andrew.cmu.edu}
\author{Ozan K. Tonguz}
\authornotemark[1]
\email{tonguz@andrew.cmu.edu}
\affiliation{%
  \institution{Carnegie Mellon University}
  \city{Pittsburgh}
  \state{Pennsylvania} 
  \country{USA}
}








\renewcommand{\shortauthors}{Almansour and Tonguz}

\begin{abstract}
Machine Learning (ML) models are extensively used in various applications due to their significant advantages over traditional learning methods. However, the developed ML models often underperform when deployed in the real world due to the well-known distribution shift problem. This problem can lead to a catastrophic outcomes when these decision-making systems have to operate in high-risk applications. Many researchers have previously studied this problem in ML, known as distribution shift problem, using statistical techniques (such as Kullback-Leibler, Kolmogorov-Smirnov Test, Wasserstein distance, etc.) to quantify the distribution shift. In this letter, we show that using Characteristic Function (CF) as a frequency domain approach is a powerful alternative for measuring the distribution shift in high-dimensional space and for domain adaptation.
\end{abstract}




\maketitle


\section{Introduction}
In recent years, Machine Learning (ML) models have been widely used due to their outstanding performance in various applications. However, not every developed model can be used in practice with confidence when they encounter data obtained from different environments or setups that are significantly different from what was used during training. Several previous studies have shown that ML models’ performance degrades, often severely, when the test data follows a different probability distribution compared to the probability distribution of the training dataset. This is known as the the distribution shift problem in ML systems.

In this work, we propose to use Characteristic Function (CF) for measuring the distribution shift and domain adaptation. Our idea is motivated by a key concept in signal processing which is: working with signals using their frequency domain representation may, in some cases, significantly simplify many computational problems compared to doing processing in the spatial or time domain. Most of the approaches reported in the literature address the distribution shift problem using features in the spatial domain, without considering how other data representations can help mitigating this problem \cite{surv1}, \cite{surv2}. These approaches typically measure the distance between the Probability Density Functions (PDFs) of the training data and the test data. However, measuring the distance between the distributions or PDFs in high-dimensional spaces is notoriously difficult because estimating N-dimensional probability distributions is hard. The proposed CF approach circumvents this difficulty. It is well-known in probability theory that using the CF of a random variable is a powerful approach for many operations as it always exist for real-valued random variables and it corresponds to the Fourier Transform of the PDF of a random variable. Our main contribution is presenting a new approach for measuring the gap between domains when using the CF since it shows that these gaps can be clearly visualized, analyzed, and mitigated.


\section{Problem Formulation} \label{probform}

\textbf{Characteristic Function:} Processing of signals, particularly images, is usually done in the spatial domain. However, for certain applications, doing processing in the frequency domain can significantly simplify analysis. The CFs are essentially Fourier Transforms of PDFs, which provide an alternative way for representing the distribution of a random variable \cite{probabilityTheory}, \cite{generativeCF} and a powerful tool to analyze them. Given a random variable \( X \), its CF is mathematically defined by Equation \ref{full_cf} where \(j\) and \(W\) denote the imaginary number and the projected frequency vector, respectively:

\begin{equation}
\phi_X(w) = \mathbb{E}\left[ e^{j W^\top X} \right] = \mathbb{E}\left[ e^{j \sum_{k=1}^{n} w_k x_k} \right]
\label{full_cf}
\end{equation} 

\textbf{Measuring the Domain Shift:} Considering the use of the CF requires first answering the following question:\textit{ does using Characteristic Functions (CFs) help in measuring the distribution shift when we have data drawn from different distributions or sample populations?} To investigate the usefulness of this approach, we use the PACS dataset \cite{pacs} which is a frequently used benchmark that consists of samples from different domains. In Appendix \ref{PACS_Samples}, we show samples from this benchmark in Figure \ref{domains}. For clarity, we did the analysis by focusing on each class separately. We took batches of images from different domains where all these images are for the same object. Then, we extract features from the images to avoid working with them at the pixel level. Since we did not know the exact PDF of each image, we used the Empirical Characteristic Function (ECF):  

\begin{equation}
\hat{\phi}_X(W) = \frac{1}{n} \sum_{k=1}^{n} e^{j W^\top X^{(k)}}
\label{emp_cf}
\end{equation}

\begin{figure}[htbp]
  \centering
  \begin{minipage}{0.98\textwidth}
    \centering
    \begin{subfigure}[b]{0.30\linewidth}
      \centering
      \includegraphics[width=\linewidth]{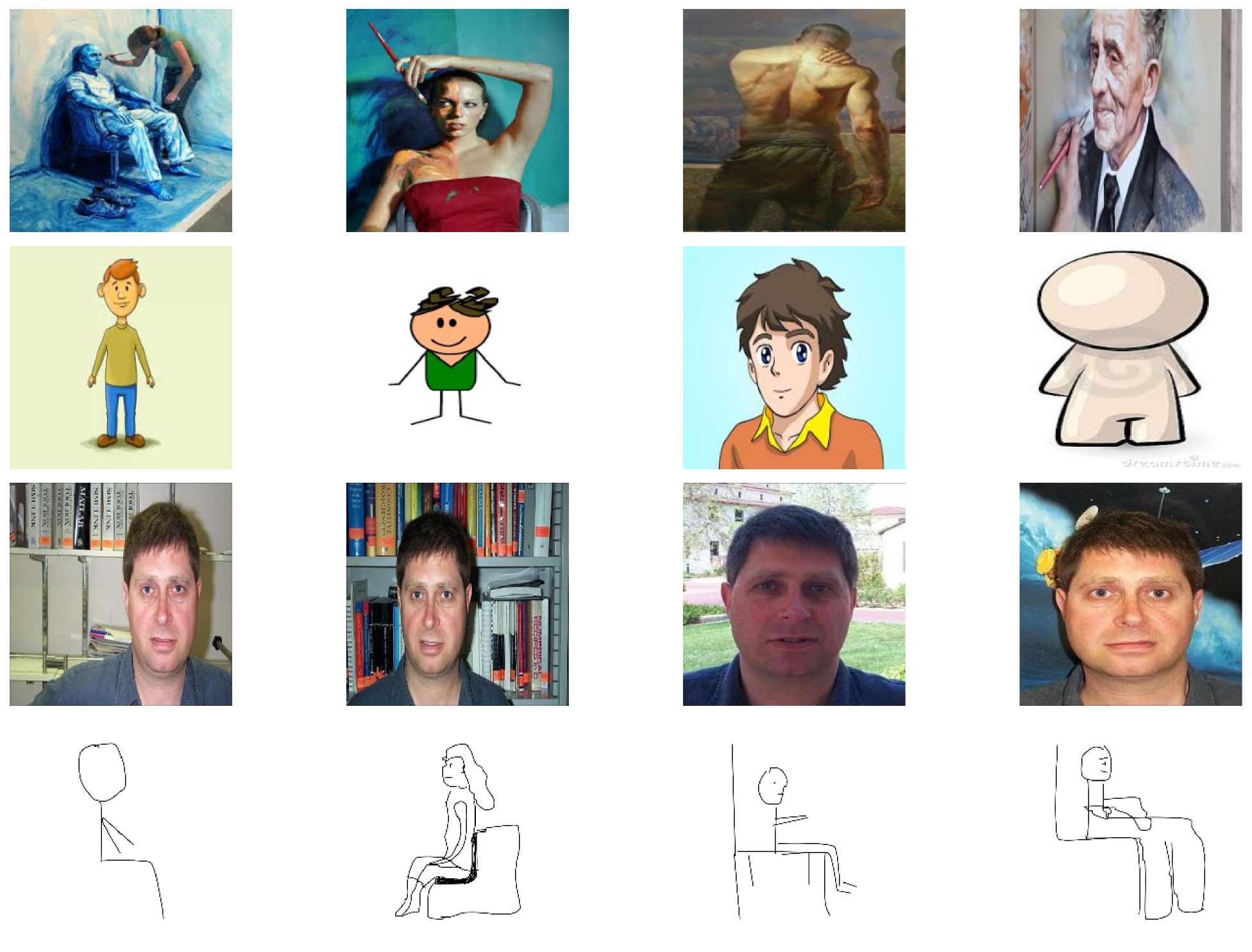}
      \caption{ }
      \label{fig:a}
    \end{subfigure}
    \hfill
    \begin{subfigure}[b]{0.33\linewidth}
      \centering
      \includegraphics[width=\linewidth]{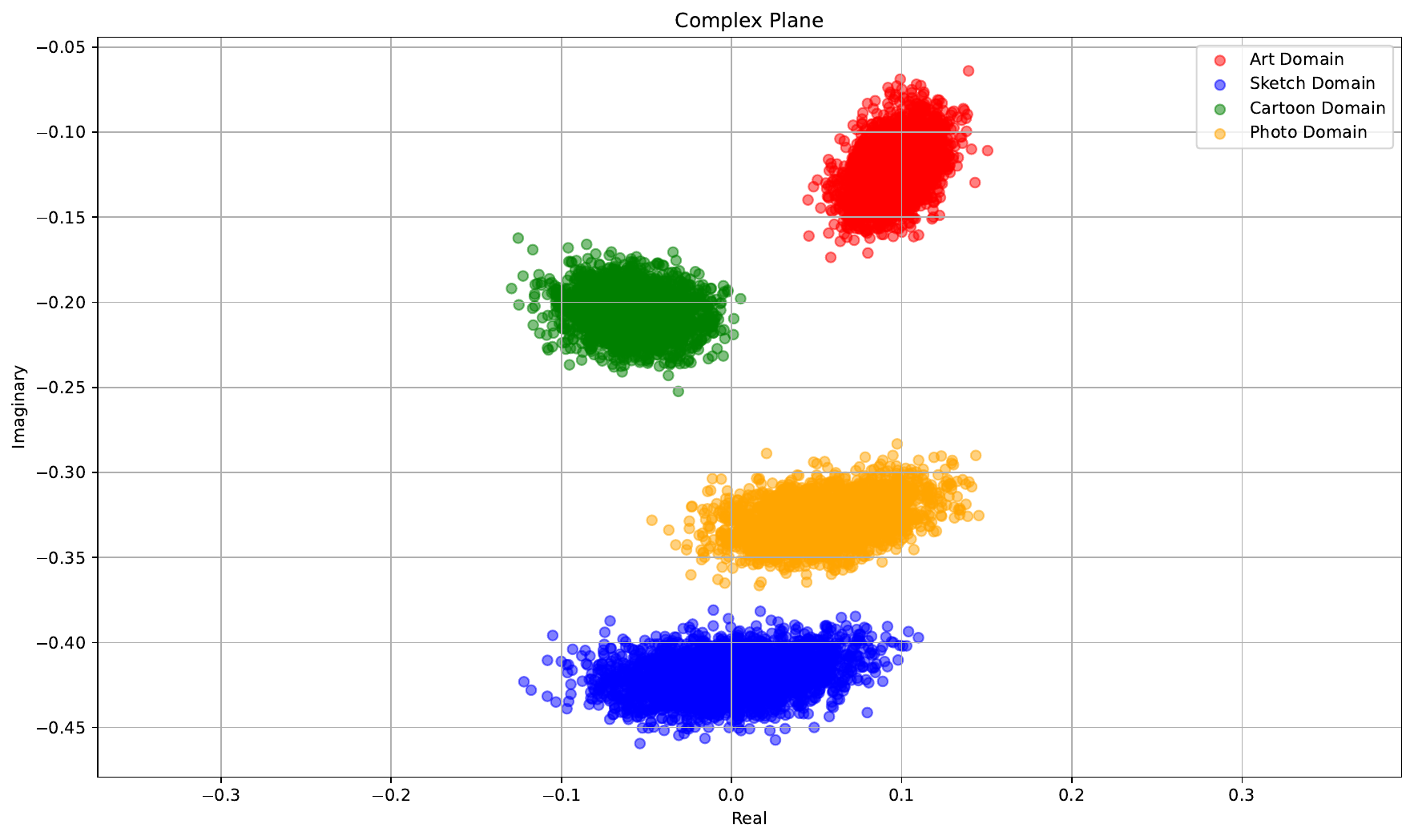}
      \caption{ }
      \label{fig:b}
    \end{subfigure}
    \hfill
    \begin{subfigure}[b]{0.33\linewidth}
      \centering
      \includegraphics[width=\linewidth]{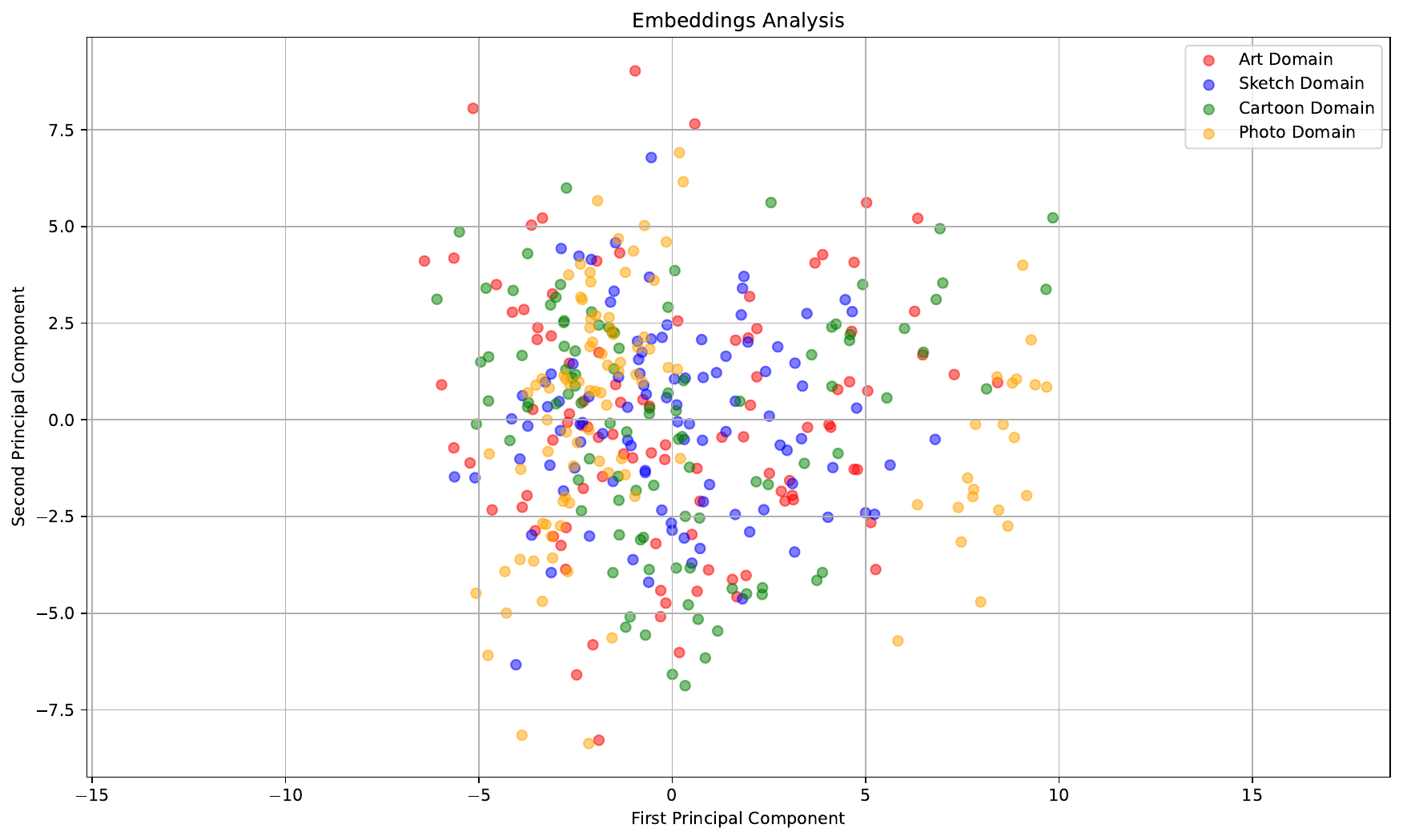}
      \caption{ }
      \label{personc}
    \end{subfigure}
    \caption{Samples from the Person class in (a) where (b) and (c) presents the Complex Plane showing domain gaps in the frequency domain and first two principal components of embeddings in the spatial domain, respectively. Observe how the gap or domain shift between different domains is visually depicted whereas it is difficult to visualize this in (c).}
    \label{personPACS}
  \end{minipage}
\end{figure}

For extracting features of images, we used a ResNet50 \cite{resnet} model pretrained on ImageNet \cite{imagener} weights as a general visual feature extractor. As an example, Figure \ref{personPACS} illustrates that the CF can result in a better understanding of the shift in the dataset as calculating it and then visualizing the results provided a better understanding of the underlying distribution shift between domains. On the other hand, the distribution shift might not be discernible if embeddings are directly analyzed in the spatial domain, as shown in Figure \ref{personc}. The latter figure shows the first two principal components of the embeddings using the Principal Component Analysis (PCA) \cite{pca}.

\section{Method} \label{method}

\textbf{Notation and Problem Setup:} Distribution shift arises when developing machine learning models using data from source/training domains and then testing them on unseen target/testing domains. Mathematically, $D = \{D_1, \ldots, D_m\}$ is our set of $m$ domains, while source domains is a subset of these different domains where $D_S = \{(X_i^S, Y_i^S)\}_{i=1}^{N_S}$ denotes the set of $X$ samples and the corresponding associated $Y$ labels for the $N_S$ samples while the target domains are denoted as $D_T = \{(X_i^T, Y_i^T)\}_{i=1}^{N_T}$, which is a subset domain comprising $N_T$ samples. The objective is to learn a model in a way that mitigates the domain discrepancy between these two subsets such that when deployed and operated on data in unseen domains, the performance does not severely degrade.

\textbf{Distribution Matching:} The common and most direct approach to train a machine learning model is Expected Risk Minimization (ERM) \cite{ERM} which minimizes the expected risks across all samples and across all different training domains, as formulated in Nguyen et al. \cite{rdm} and shown in Equation \ref{erm}, where $D$ is the set of all different domains and $\ell(.,.)$ can be any well-defined loss function. However, when considering only this setup, it leads to poor generalization results since models tend to capture more domain-specific features due to the lack of any training restrictions in this learning scheme \cite{rdm}. While there are several methods in the literature to tackle this problem, the distribution matching-based methods are the most direct approaches. Instead of solely utilizing $\ell_{ERM}$ for the learning task, an additional term can formulate a total loss function for a more robust learned model. The additive term provides more insights into the domain variations and hence minimizes such discrepancy. Then, Equation \ref{total_loss_general} becomes the new objective function to optimize as the $\ell_{shift}$ is the new additional term for capturing the domain variations and for restricting the training process. Here, $\lambda \geq 0$ is a regularizer balancing between reducing the two terms.

\begin{equation}
\ell_{ERM} = \mathbb{E}_{d \sim D}\ \mathbb{E}_{(x_d, y_d) \sim D_d} \left[ \ell(f(x_d), y_d) \right]  
\label{erm}
\end{equation} 

\begin{equation}
    \ell_{total} = \ell_{ERM} + \lambda \ell_{shift} 
    \label{total_loss_general}
\end{equation}

\textbf{Characteristic Function Loss:} The proposed idea is to utilize the CF of images’ embeddings from different domains as a tool to minimize the discrepancy between the domains' features. The intuition behind this idea is: since the CF always exists (as long as the PDF of the random variable exists) and uniquely defines a random variable, it might be an effective approach to use for measuring and minimizing the distribution shift \cite{generativeCF}. Considering that each image is a random variable, and random variables from the same domain should have similar CFs, the main idea is to match embeddings of images from different domains to minimize the gap between domains by constructing a random vector of image embeddings as \( X = (X_1, \ldots, X_n) \), where the RVs \( X_i \) are independent. Since we do not know the exact PDF of each image, we use the ECF presented in Equation \ref{emp_cf} and then use the computed values for the extracted features from different domains as our added distribution shift loss to the total loss, as shown in Equations \ref{l2_cf} and \ref{total_loss}.

\begin{equation}
\ell_{CFL} = \frac{1}{N} \sum_{N=1}^{W} \left( \phi_{S,N}(W) - \phi_{T,N}(W) \right)^2
\label{l2_cf}
\end{equation}

\begin{equation}
    \ell_{total} = \ell_{ERM} + \lambda \ell_{CFL} 
    \label{total_loss}
    \end{equation}


\section{Results and Discussion} \label{discussion}

\begin{figure}[htbp]
  \centering
  \begin{minipage}{0.98\textwidth}
    \centering
    \begin{subfigure}[b]{0.30\linewidth}
      \centering
      \includegraphics[width=\linewidth]{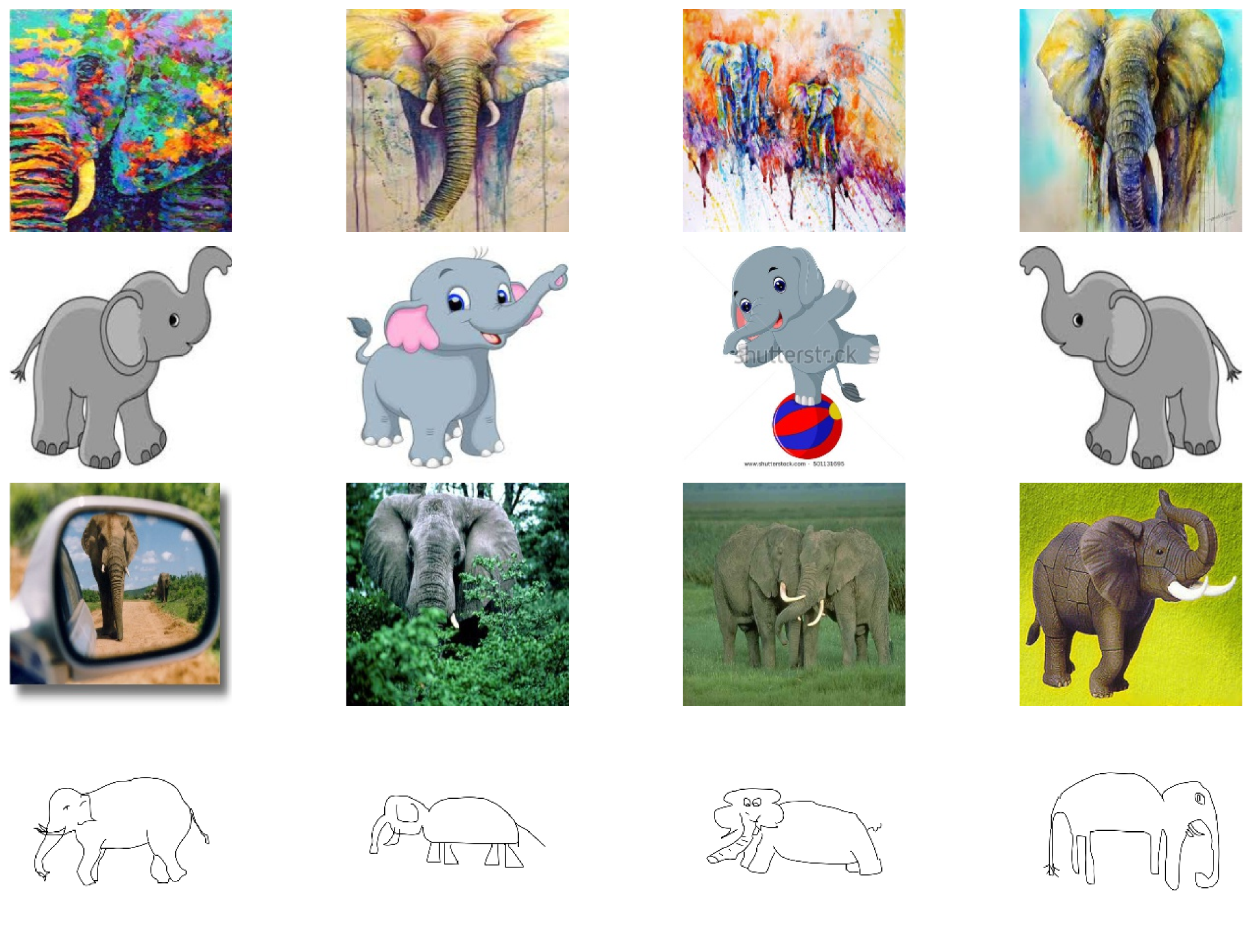}
      \caption{ }
      \label{fig:a}
    \end{subfigure}
    \hfill
    \begin{subfigure}[b]{0.33\linewidth}
      \centering
      \includegraphics[width=\linewidth]{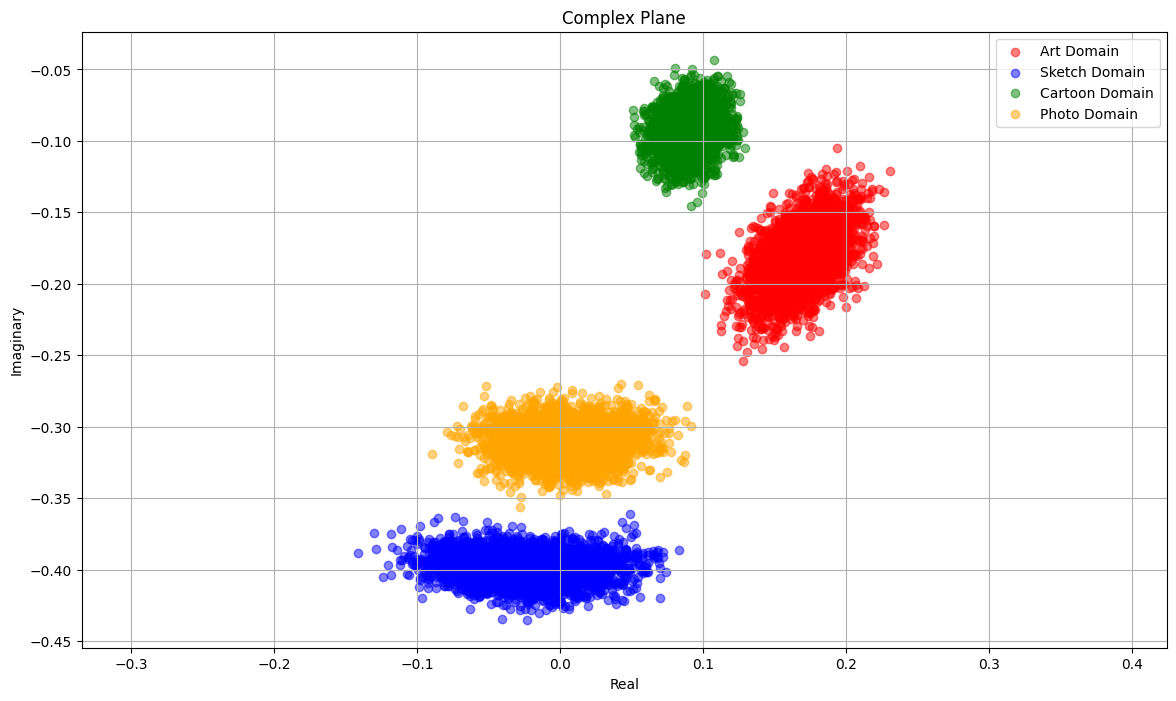}
      \caption{ }
      \label{fig:b}
    \end{subfigure}
    \hfill
    \begin{subfigure}[b]{0.33\linewidth}
      \centering
      \includegraphics[width=\linewidth]{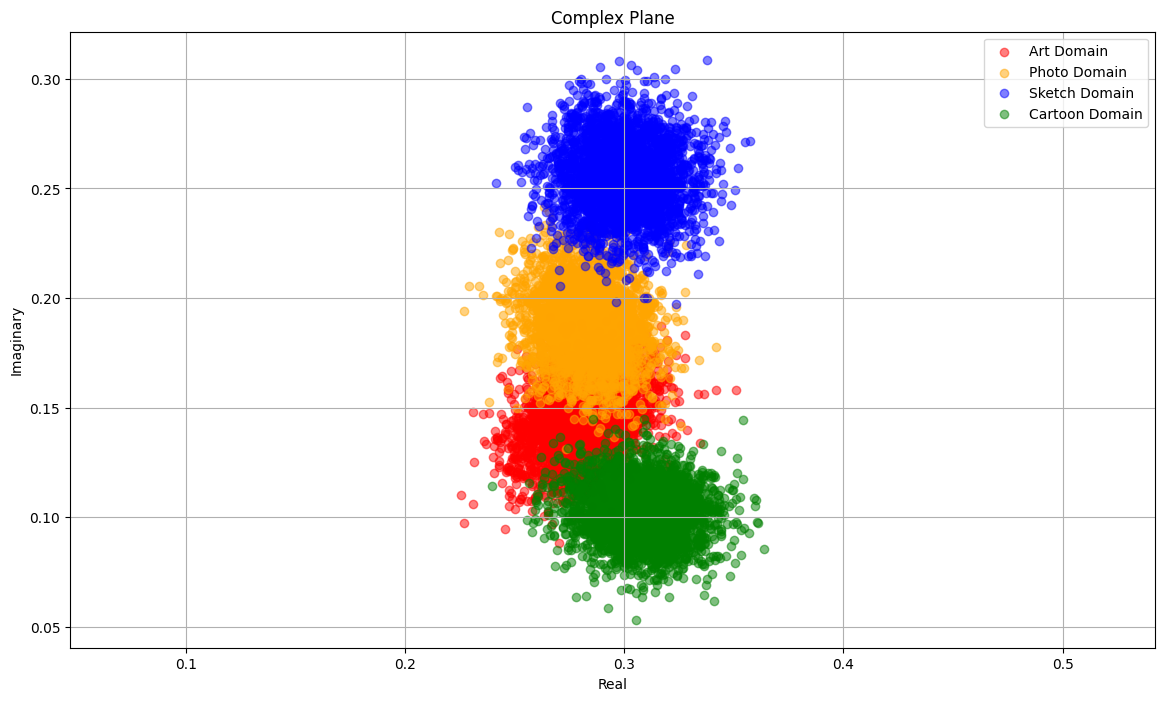}
      \caption{ }
      \label{}
    \end{subfigure}
    \caption{Samples from the Elephant class in (a) where (b) and (c) presents the Complex Plane of the distribution of domains for the backbone pretrained model and after the training using our approach, respectively. Using the CF approach can minimize such divergence between domains resulting in a model that performs well in the deployment environment with unforeseen domains.}
    \label{results}
  \end{minipage}
\end{figure}

\textbf{Distributions Alignment:} As shown in Figure \ref{results} and quantified in Table \ref{distances}, starting with a dataset consisting of very different domains, using the CF for training the model results in a model that minimizes such discrepancy. The Figure and the Table below present how the trained model on Cartoon and Sketch domains resulted in a minimization of the gap between them. Observe that the CF approach used is able to project domains that were not used during the training (Photo and and Art Painting domains) closely to those used for training the model. Hence, this shows that the model is being guided during the training to focus more on domain invariant features. The experimental setup details for this analysis is discussed in Appendix \ref{setup}. Furthermore, this approach can help in assessing the certainty of the model prediction. Considering the data at the inference stage, such feature representation can facilitate measuring by how far the input sample is from the location of the clustered domains.

\begin{table}[h]
  \centering
  \caption{Distances between domains for the pretrained model and after using the CF. Green cells with $\downarrow$ indicate decreased distances between domains. The used distance metric is computed using Equation \ref{l2_cf}.}
  \label{distances}
  \adjustbox{max width=\columnwidth}{
  \begin{tabular}{lcccccccc}
    \toprule
    \textbf{Setups $\Rightarrow$} & \multicolumn{4}{c}{\textbf{Before Training}} & \multicolumn{4}{c}{\textbf{After Training}} \\
    \cmidrule(lr){2-5} \cmidrule(lr){6-9}
    \textbf{Domains} & Art & Sketch & Cartoon & Photo & Art & Sketch & Cartoon & Photo \\
    \midrule
    Art     & -- & 0.085 & 0.014 & 0.044 & -- & \cellcolor{green!15}0.013 ($\downarrow$84.7\%) & \cellcolor{green!15}0.003 ($\downarrow$78.5\%) & \cellcolor{green!15}0.002 ($\downarrow$95.4\%) \\
    Sketch  & 0.085 & -- & 0.107 & 0.009 & \cellcolor{green!15}0.013 ($\downarrow$84.7\%) & -- & \cellcolor{green!15}0.024 ($\downarrow$77.5\%) & \cellcolor{green!15}0.005 ($\downarrow$44.4\%) \\
    Cartoon & 0.014 & 0.107 & -- & 0.055 & \cellcolor{green!15}0.003 ($\downarrow$78.5\%) & \cellcolor{green!15}0.024 ($\downarrow$77.5\%) & -- & \cellcolor{green!15}0.008 ($\downarrow$85.4\%) \\
    Photo   & 0.044 & 0.009 & 0.055 & -- & \cellcolor{green!15}0.002 ($\downarrow$95.4\%) & \cellcolor{green!15}0.005 ($\downarrow$44.4\%) & \cellcolor{green!15}0.008 ($\downarrow$85.4\%) & -- \\
    \bottomrule
  \end{tabular}}
\end{table}


\section{Conclusion} \label{conclusion}

In this letter, we have proposed using the Characteristic Function (CF) approach to measure and analyze the distribution shift and minimize it for a given dataset. Our method utilizes the CF as a frequency domain approach to quantify the distribution shift. A frequency domain approach is a powerful alternative to the existing distribution matching based methods that minimize the distances between domains. Moreover, since the CF is the Fourier Transform of the PDF of a random variable, it avoids the estimation of probability distributions in high-dimensional space which is a difficult problem and also provides an alternative way to measure the distribution shift. This, in turn, helps domain adaptation by reducing the distribution shift between domains.


\bibliographystyle{ACM-Reference-Format}
\bibliography{references}


\appendix


\section{Examples Samples from PACS Dataset} \label{PACS_Samples}
\begin{figure} [H]
\centering
\includegraphics[width=14cm]{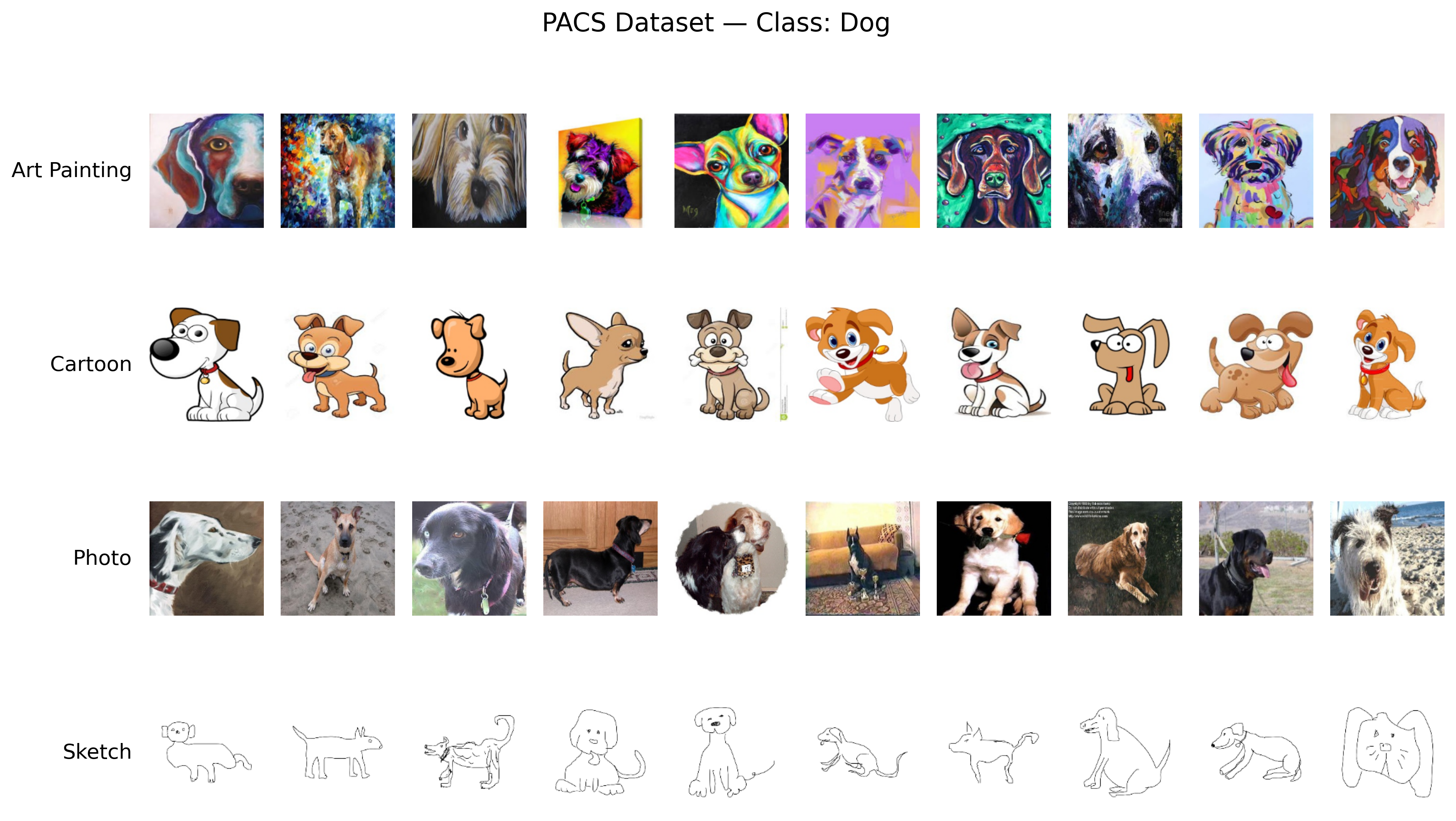} 
\caption{Example which demonstrate different manifested distribution shift scenarios in the PACS \cite{pacs} dataset for the object class Dog.}
\label{domains}
\end{figure}  


\section{Implementation Details} \label{setup}
Our analysis pipeline is built on the code base of the DeepDG \cite{deepdg} which is a well-designed tool-kit for distribution shift research. Our idea was implemented using this framework; the main parameters modified are the learning rates and our loss function regularizer, $\lambda$, with a value of 0.001 and 0.1, respectively. The followed model selection algorithm is the model at the last training epoch. For the training and evaluation, we utilized a widely used challenging benchmark, namely the PACS \cite{pacs} dataset. This benchmark consists of around 9K images and 7 classes, where samples are distributed among four different domains: Cartoon, Sketch, Photo and Art Painting. In our analysis, we considered only the Cartoon and Sketch as our source (training) and target (validation) domains, respectively, for this study while the others were left as unseen domains for future analysis.

\end{document}